\documentclass{ecai2010}
\usepackage{epsfig}
\usepackage{epsfig,latexsym, amssymb, amsmath}
\usepackage{lscape}

\newcommand\T{\rule{0pt}{2.6ex}}

\newcommand{\hide}[1]{}

\begin{document}

\title{Adaptive Branching for Constraint Satisfaction Problems}

\author{Thanasis Balafoutis \and Kostas Stergiou\institute{Department of Information \& Communication Systems Engineering University of the Aegean, Greece, email: \{abalafoutis,konsterg\}@aegean.gr} }

\maketitle
\bibliographystyle{plain}

\begin{abstract}

The two standard branching schemes for CSPs are $d$-way and 2-way branching. Although it has been shown that in theory the latter can be exponentially more effective than the former, there is a lack of empirical evidence showing such differences. To investigate this, we initially make an experimental comparison of
the two branching schemes over a wide range of benchmarks. Experimental results verify the theoretical gap between $d$-way and 2-way branching as we move from a simple variable ordering heuristic like $dom$ to more sophisticated ones like $dom/ddeg$. However, perhaps surprisingly, experiments also show that when state-of-the-art variable ordering heuristics like $dom/wdeg$ are used then $d$-way can be clearly more efficient than 2-way branching in many cases. Motivated by this observation, we develop two generic heuristics that can be applied at certain points during search to decide whether 2-way branching or a restricted version of 2-way branching, which is close to $d$-way branching, will be followed. The application of these heuristics results in an adaptive branching scheme. Experiments with instantiations
of the two generic heuristics confirm that search with adaptive
branching outperforms search with a fixed branching scheme
on a wide range of problems.

\end{abstract}

\section{INTRODUCTION}


Most complete algorithms for CSPs are based on exhaustive backtracking search interleaved with constraint propagation. Search is typically guided by variable and value ordering heuristics and makes use of either a ``$d$-way'' or a ``2-way'' branching scheme. In the former case, after a variable $x$ with domain $\{a_1,\ldots, a_d\}$ is chosen, $d$ branches are built, each one corresponding to one of the $d$ possible value assignments of $x$. In the latter case, after a variable $x$ is chosen, its values are assigned through a sequence of binary choices. The first choice point creates two branches, corresponding to the assignment of $a_1$ to $x$ (left branch) and the removal of $a_1$ from the domain of $x$ (right branch). 2-way branching was described by Freuder and Sabin within the MAC algorithm \cite{sabin97} and in theory it can achieve exponential savings in search effort compared to $d$-way branching \cite{mitchell05}. Indeed, 2-way is the standard branching scheme of most constraint solvers.

Despite the theoretical result of \cite{mitchell05}, the few experimental studies comparing 2-way and $d$-way branching have not displayed significant differences between them. Park showed that 2-way and $d$-way 
display very similar performance when the smallest domain variable ordering heuristic (VOH) is used \cite{park04}, while Smith and Sturdy showed that 2-way outperforms $d$-way when searching for all solutions, albeit not considerably (an average speed-up of 30$\%$ was reported) \cite{smith05}. The reason for this ``discrepancy'' between theory and experiments is that the experimental studies have considered a ``restricted'' form of 2-way branching where the variable branched on after the successful propagation of a value removal is always the current variable. This is also the default branching scheme of most constraint solvers. Importantly, ``full'' 2-way branching as described in \cite{sabin97}, allows for a different variable to be chosen.

In this paper we first make a detailed
experimental comparison between 2-way branching, in both its restricted and full versions, and $d$-way branching, under a variety of different VOHs. Results show that the $d$-way and restricted 2-way branching schemes are closely matched across the different VOHs, with $d$-way being slightly more cost effective. However, confirming the theoretical results, exponential differences in favor of full 2-way branching are observed as soon as we move from a simple heuristic like smallest domain ({\em dom}) to more sophisticated ones like domain over dynamic degree ({\em dom/ddeg}). Perhaps surprisingly, when state-of-the-art conflict-driven heuristics, like {\em dom/wdeg}, are used, significant differences in favor of $d$-way (and restricted 2-way) are also observed. This is because in some cases the VOH mistakenly chooses to branch on a variable other than the current one after the successful propagation of a value removal. This can divert search away from a hard part of the search space, resulting in increased search effort.

Motivated by this observation, we develop two generic heuristics that can be applied at successful right branches once the VOH chooses to branch on a variable other than the current one. At this point the heuristics are used to decide whether the advice of the VOH will be followed or not. The application of these heuristics results in an adaptive branching scheme that dynamically switches between 2-way branching and its restricted version (which is close to $d$-way branching). Both of our heuristics can be used in tandem with any backtracking search algorithm and VOH. The first heuristic is based on measuring the difference between the scores that the VOH assigns to its selected variable and the current variable. The VOH is followed only if the difference is sufficiently large. As a downside, this heuristic requires some tuning to optimize its performance. The second heuristic is based on the use of a secondary advisor to decide if the VOH will be followed, and it does not require any tuning.

Experiments with instantiations of the two generic heuristics confirm that search with adaptive branching outperforms search with a fixed branching scheme on a wide range of problems. Interestingly, in many cases the gains offered by full 2-way branching are also obtained by the adaptive branching methods with only very few decisions following the VOH when it suggests to move away from the current variable at successful right branches.

The rest of the paper is organized as follows. Section~\ref{section-background} gives necessary background. In Section~\ref{section-comparison} we compare the three branching schemes detailed above on a variety of CSP instances. Section~\ref{section-heuristics} presents the proposed heuristics for adaptive branching. In Section~\ref{section-experiments} we demonstrate the efficacy of the heuristics through an experimental study. Finally, in Section~\ref{section-conclusions} we conclude.

\section{BACKGROUND}

\label{section-background}


A \emph{Constraint Satisfaction Problem} (CSP) is a tuple
(\emph{X, D, C}), where \emph{X} is a set containing \emph{n}
variables \{\emph{$x_1, x_2,..., x_n$}\}; \emph{D} is a set of
domains \{\emph{$D(x_1)$, $D(x_2)$,..., $D(x_n)$}\} for those
variables, with each $D(x_i)$ consisting of the possible values
which $x_i$ may take; and \emph{C} is a set of constraints
between variables in subsets of
\emph{X}. Each constraint $c_i \in C$ expresses a relation defining which
variable assignment combinations are allowed for the variables in
the scope of the constraint.
The {\em degree} of a variable $x$ is the number of constraints involving $x$, and the {\em dynamic degree} of (an unassigned variable) $x$ is the number of constraints involving $x$ and at least one other unassigned variable.

Complete search algorithms for CSPs are typically based on backtracking depth-first search where branching decisions (e.g. variable assignments) are interleaved with constraint propagation. Search is guided by variable ordering heuristics (VOHs) and value ordering heuristics. The classic VOH {\em smallest domain} ({\em dom}) selects the variable with minimum domain size \cite{haral80}. Other, more sophisticated, heuristics include {\em dom/deg} and {\em dom/ddeg} \cite{bessiere96}, which select the variable with minimum ratio of domain size over degree (resp. dynamic degree).

One of the most efficient general purpose VOHs that have been proposed is {\em dom/wdeg} \cite{bhls04}. This heuristic assigns a weight to each constraint, initially set to one. Each time a constraint causes a conflict, i.e. a domain wipeout (DWO), its weight is incremented by one. Each variable is associated with a {\em weighted degree}, which is the sum of the weights over all constraints involving the variable and at least another unassigned variable. The {\em dom/wdeg} heuristic chooses the variable with minimum ratio of current domain size to weighted degree. This heuristic is among the most efficient, if not {\em the} most efficient, general-purpose heuristics for CSPs. Grimes and Wallace proposed alternative conflict-driven heuristics that consider value deletions as the basic conflicts associated with constraint weights \cite{grim07}. The {\em alldel} heuristic increments the weight of a constraint each time it causes one or more value deletions from a domain. The efficacy of all the proposed conflict-directed heuristics is due to their ability to learn though conflicts (either DWOs or value deletions) encountered during search. As a result they can guide search towards hard parts of the problem earlier.

\begin{figure}[htb]
\begin{tabular}{c}
\includegraphics{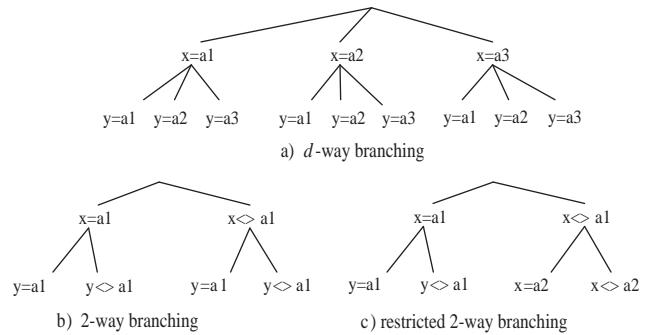}
\end{tabular}
\vspace{-5mm}
\caption{Examples of search trees for the three branching schemes.}
\label{fig-branching}
\end{figure}

A CSP search algorithm is usually implemented using either a \emph{d-way} or a \emph{2-way} branching scheme\footnote{Domain splitting is also used, but it is not considered here.}. The former works as follows. After a variable $x$ with domain $D(x)=\{a_1, a_2,..., a_d\}$ is selected, $d$ branches are created, each one corresponding to a value assignment of $x$. In the first branch, value $a_1$ is assigned to $x$ and constraint propagation is triggered. If this branch fails, $a_1$ is removed from $D(x)$. Then the assignment of $a_2$ to $x$ is made (second branch), and so on. If all $d$ branches fail then the algorithm backtracks. An example of a search tree explored with $d$-way branching is shown in Figure~\ref{fig-branching}a. In \emph{2-way} branching, after a variable $x$ and a value $a_i \in D(x)$ are selected, two branches are created. In the left branch $a_i$ is assigned to $x$, or in other words the constraint $x$=$a_i$ is added to the problem and is propagated. In the right branch the constraint $x\neq a_i$ is added to the problem and is propagated. If both branches fail then the algorithm backtracks. Figure~\ref{fig-branching}b shows a search tree explored with 2-way branching.

There are two differences between these branching schemes:
\begin{itemize}
\item In 2-way branching, if the branch assigning a value $a_i$ to a variable $x$ fails then the removal of $a_i$ from $D(x)$ is immediately propagated. Instead, $d$-way branching tries the next available value $a_j$ of $D(x)$. Note that the propagation of $a_j$ subsumes the propagation of $a_i$'s removal.

\item In 2-way branching, after a failed branch corresponding to an assignment $x$=$a_i$, and assuming the removal of $a_i$ from $D(x)$ is then propagated successfully, the algorithm can choose to branch on any variable (not necessarily $x$), according to the VOH (e.g. Figure~\ref{fig-branching}b). In $d$-way branching the algorithm has to choose the next available value for variable $x$ after $x$=$a_i$ fails.
\end{itemize}

In between these two schemes is the version of 2-way branching used in \cite{smith05} where the algorithm is forced to assign $x$ with its next value after the successful propagation of $a_i$'s removal from $D(x)$.
In the following we call this {\em restricted 2-way branching}. Figure~\ref{fig-branching}c shows a search tree explored with restricted 2-way branching.

\section{COMPARISON OF BRANCHING SCHEMES}

\label{section-comparison}

The search algorithm used in all the experiments presented hereafter is the commonly used maintaining (generalized) arc consistency (MAC) algorithm \cite{sabin94,bessiere96}. We have experimented with four VOHs: {\em dom}, {\em dom/ddeg}, {\em dom/wdeg}, and {\em dom/alldel}. Benchmarks
are taken from the web page of C. Lecoutre, where details about them can be
found. The following classes of binary and non-binary problems have been used: radio links frequency assignment (RLFAP), driver, ehi, geom, graph coloring, quasigroup completion, chessboard coloration, all-interval series, golomb ruler and random instances from the RB model (forced to be satisfiable).

Results from the use of {\em dom} with the three branching schemes showed that under this heuristic 2-way branching essentially emulates $d$-way branching and the three branching schemes are closely matched in terms of run times (with $d$-way being slightly better). This is in accordance with previous results \cite{park04}.

\begin{table}
\caption{CPU times (t) in secs. and nodes (n) for the three branching schemes using the VOH {\em dom/ddeg}.}
\centering
\begin{scriptsize}
\begin{tabular}{|c|c|c|c|c|}
\hline
Instance \T&  & $2-way$ & $restricted$ & $d-way$ \\
 &  &  & $2-way$ &  \\
\hline 
scen-11 \T& t & \textbf{7.1} & 508 & 372\\
$(sat)$ & n & 1379 & 68597 & 45862 \\ \hline
scen-11-f11 \T& t & \textbf{8.1} & $>1h$ & $>1h$ \\
$(unsat)$ & n & 2716 & - & - \\ \hline
driverlogw-09 \T& t & \textbf{205} & 1197 & 1207 \\
$(sat)$ & n & 75033 & 0.42M & 0.16M \\ \hline
ehi-85-297-0 \T& t & \textbf{62.3} & 2509 & 2481 \\
$(unsat)$ & n & 37246 & 2.5M & 0.9M \\ \hline
ehi-85-297-2 \T& t & \textbf{41.8} & $>1h$ & $>1h$ \\
$(unsat)$ & n & 20733 & - & - \\ \hline
ash958GPIA-4 \T& t & \textbf{5.1} & $>1h$ & $>1h$ \\
$(sat)$ & n & 2712 & - & - \\ \hline
cc-12-12-2 \T& t & \textbf{58.3} & 978 & 959 \\
$(unsat)$ & n & 36994 & 0.56M & 0.27M \\ \hline
\end{tabular}
\end{scriptsize}
\label{table:ddeg} 
\end{table}

In Table~\ref{table:ddeg} we give indicative results from the use of the {\em dom/ddeg} heuristic. We display CPU times as well as nodes. A node in 2-way branching can be a value assignment or a value removal, while in $d$-way branching it can only be a value assignment. Hence, they cannot be compared directly. However, nodes give an accurate measure of the difference in search effort between 2-way and restricted 2-way. Our experiments confirm the theoretical results of \cite{mitchell05} as we can observe huge differences in favor of 2-way branching. Restricted 2-way and $d$-way display similar performance.

Table~\ref{table:wdeg} compares the three branching schemes when using the conflict-driven VOH \emph{dom/wdeg}. As this heuristic can learn from failures (DWOs) 
encountered during search, and thus make more informed choices, it significantly outperforms {\em dom/ddeg} \cite{bhls04}. Comparing these results with the results of Table~\ref{table:ddeg}, we can notice two differences: First, there are instances where 2-way branching is less efficient, sometimes considerably, compared to restricted 2-way and $d$-way.
Second, in instances where 2-way branching dominates, the differences are not as striking as in Table~\ref{table:ddeg}, albeit they can still be considerable. Quantitatively similar results were observed when using the conflict-driven VOH {\em dom/alldel}.

\begin{table}
\caption{CPU times (t) in secs. and nodes (n) for the three branching schemes using the VOH {\em dom/wdeg}.}
\centering
\begin{scriptsize}
\begin{tabular}{|c|c|c|c|c|}
\hline
Instance \T&  & $2-way$ & $restricted$ & $d-way$ \\
 &  &  & $2-way$ &  \\
\hline 
series-12 \T& t & \textbf{27.3} & 145.7 & 158.5\\
$(sat)$ & n & 33876 & 245787 & 246379 \\ \hline
series-13 \T& t & \textbf{148.4} & 1132 & 1057 \\
$(sat)$ & n & 150145 & 1.6M & 1.3M \\ \hline
scen11 \T& t & \textbf{12.6} & 45.1 & 34.2 \\
$(sat)$ & n & 2414 & 6002 & 4398 \\ \hline
scen2-f25 \T& t & \textbf{42.3} & 183.8 & 137.9 \\
$(unsat)$ & n & 9539 & 38185 & 50609 \\ \hline
graph9-f9 \T& t & \textbf{178.6} & 640.5 & 923 \\
$(unsat)$ & n & 66367 & 255204 & 406361 \\ \hline
ruler-25-8-a3 \T& t & \textbf{50.3} & 180.8 & 167.6 \\
$(unsat)$ & n & 1829 & 6407 & 7026 \\ \hline
frb35-17-1 \T& t & \textbf{34.2} & 135 & 114 \\
$(sat)$ & n & 11129 & 54026 & 40165 \\ \hline\hline
geo50-20-d4-75-1 \T& t & 2298 & 1242 & \textbf{1128} \\
$(sat)$ & n & 558754 & 279998 & 257225 \\ \hline
geo50-20-d4-75-2 \T& t & 122 & 28,5 & \textbf{28.3} \\
$(sat)$ & n & 48943 & 9153 & 9509 \\ \hline
haystacks-05 \T& t & 41.7 & 3.9 & \textbf{3.7} \\
$(unsat)$ & n & 1.78M & 0.18M & 0.16M \\ \hline
qcp-15-120-5 \T& t & 529 & 408 & \textbf{402} \\
$(sat)$ & n & 0.76M & 0.63M & 0.25M \\ \hline
abb313GPIA-7 \T& t & 498 & 313 & \textbf{309} \\
$(unsat)$ & n & 28388 & 17238 & 11161 \\ \hline
ruler-25-7-a3 \T& t & 12.4 & 4.1 & \textbf{2.1} \\
$(sat)$ & n & 1444 & 255 & 129 \\ \hline
frb35-17-2 \T& t & 337 & 227 & \textbf{194} \\
$(sat)$ & n & 0.13M & 0.1M & 78938 \\ \hline
\end{tabular}
\end{scriptsize}
\label{table:wdeg} 
\end{table}

In the first part of Table~\ref{table:wdeg} 
we group representative instances where 2-way branching is the best choice, while in the second part 
we show results where restricted 2-way and d-way are better. Neither 2-way nor $d$-way (and restricted 2-way) is the best choice, even for instances within the same problem class (see for example the \emph{ruler25} and \emph{frb35-17} problems in Table~\ref{table:wdeg}). As in the case of {\em dom/ddeg}, restricted 2-way branching displays a behavior very close to that of $d$-way branching.

A likely explanation for the failure of 2-way branching on some instances, compared to its restricted version, is the following. At some right branches during search, the VOH mistakenly chooses to branch on a variable other than the current one. In the case of conflict-driven VOHs, this may result in the search process moving away from a hard subproblem to another area of the search space resulting in increased search effort. To test this conjecture we have developed heuristics, presented below, that can be applied at successful right branches to decide whether the advice of the VOH will be followed or not. The use of such heuristics results in an adaptive branching scheme that dynamically switches between 2-way branching and its restricted version.

Finally, a conclusion that can be drawn from this experimental study is that the performance of different branching schemes strongly depends on the VOH used.

\section{HEURISTICS FOR ADAPTIVE BRANCHING}

\label{section-heuristics}


We now present two generic heuristics that can be used to dynamically adapt the search algorithm's branching scheme. We only consider the case where dynamic adaptation involves switching between 2-way branching and restricted 2-way branching. As detailed above, the performance of $d$-way branching is very close to that of restricted 2-way branching. The intuition behind the heuristics is twofold. First, to avoid branching on a different variable if the VOH is not ``confident enough'' about the correctness of this decision. And second, to identify ways to assist this decision by the use of secondary advisors. That is, VOHs that can complementarily be consulted to help in the decision making.

These heuristics can be applied at successful right branches. That is, when the VOH suggests to branch on another variable rather that trying the next  value of the current one. Following the above intuitions we propose two generic heuristics:

$\bf{H_{sdif{}f}(e):}$\textbf{- VOH score difference} If the current variable is $x$ and the VOH suggests to branch on a different variable $y$, we will follow this suggestion only when $|score(y) - score(x)| > e$, where $score(x)$ and $score(y)$ are the values assigned by the VOH to variables $x$ and $y$, while $e$ is a threshold value difference.

$\bf{H_{cadv}(VOH_2):}$ \textbf{- complementary advisor} If the current variable is $x$ and the VOH used by the algorithm ($VOH_1$) suggests to branch on a different variable $y$, we will follow this suggestion only when a secondary VOH ($VOH_2$) also prefers $y$ to $x$. That is, when $score_{VOH_2}(y) > score_{VOH_2}(x)$, where $score_{VOH_2}(x)$ and $score_{VOH_2}(y)$ are the heuristic values assigned by $VOH_2$ to variables $x$ and $y$\footnote{We assume that a greater score is better according to $VOH_2$.}.

Both proposed heuristics are generic, in the sense that they can be
used in tandem with any VOH and any backtracking search algorithm. However, $H_{sdiff}(e)$ requires some tuning to set the value of $e$ appropriately. In contrast, $H_{cadv}(VOH_2)$ does not require any such tuning, and can use any VOH as a secondary heuristic. The two heuristics can also be combined either conjunctively or disjunctively. In the former (resp. latter) case, the suggestion to branch on a variable different than $x$ is followed when both (resp. at least one) of the criterions for $H_{sdiff}$ and $H_{cadv}$ are satisfied. Importantly, the two proposed heuristics are lightweight, 
assuming that $VOH_2$ is not too expensive to compute.

\section{EXPERIMENTS}

\label{section-experiments}


In this section we evaluate the performance of the two heuristics on a variety of CSP instances. Before presenting the results, we discuss the tuning of heuristic $H_{sdiff}(e)$. 


\subsection{Tuning Heuristic $H_{sdiff}(e)$}


Although heuristic $H_{sdiff}(e)$ is generic and can be used together with any VOH, the optimum threshold value $e$ obviously varies among different VOHs as they may consider different metrics, such as domain sizes, constraint degrees, constraint weights, etc.
Even with a fixed VOH and within a specific problem class, the optimum threshold value $e$ may differ from instance to instance, and therefore locating it is not particularly interesting from a practical point of view. However, our experiments have demonstrated that a ``good enough'' value for $e$ that carries across different problem classes can be located for a given VOH with only a few experiments.

To find a good value for $e$ we proceeded as follows. Taking a single instance from some problem class we repeatedly solved it using the $H_{sdiff}(e)$ heuristic for branching, starting with $e$ set to 0 and gradually increasing $e$ in every repetition. Setting $e=0$ forces $H_{sdiff}(e)$ to emulate 2-way branching, while as $e$ increases, $H_{sdiff}(e)$ moves closer to restricted 2-way branching. For each run we measured the number of times when, after a failed assignment $x=a_i$ and the successful propagation of $x\neq a_i$, the VOH chose to branch on a variable $y$ other than $x$. In the following we will call this measure {\em variable changes} $(vc)$. By definition, such a situation does not occur with restricted 2-way branching, hence with this branching scheme $vc$ is always 0.

\begin{figure}
  \begin{tabular}{cc}
    \includegraphics[height=1.6in]{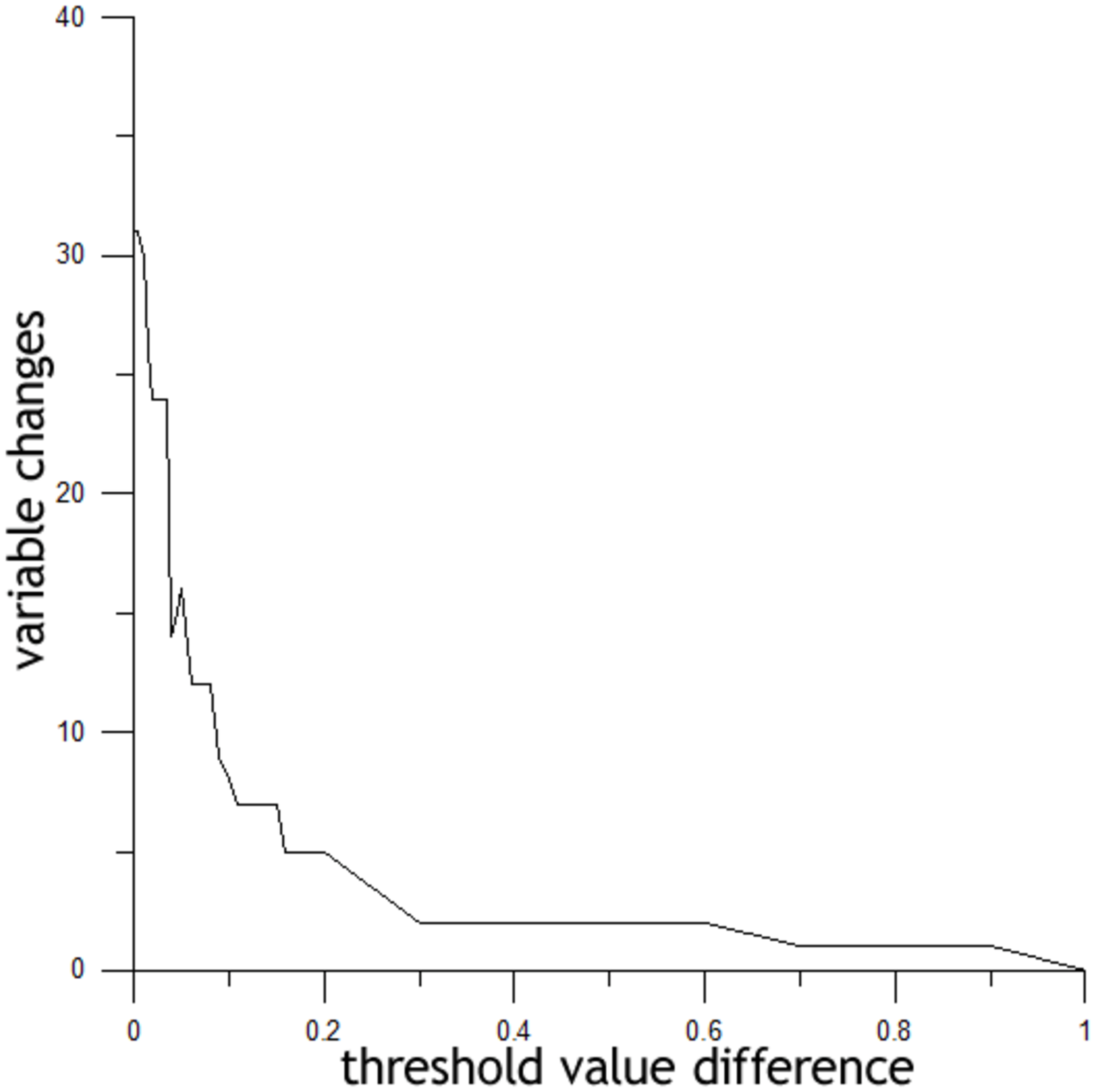} &
    \includegraphics[height=1.6in]{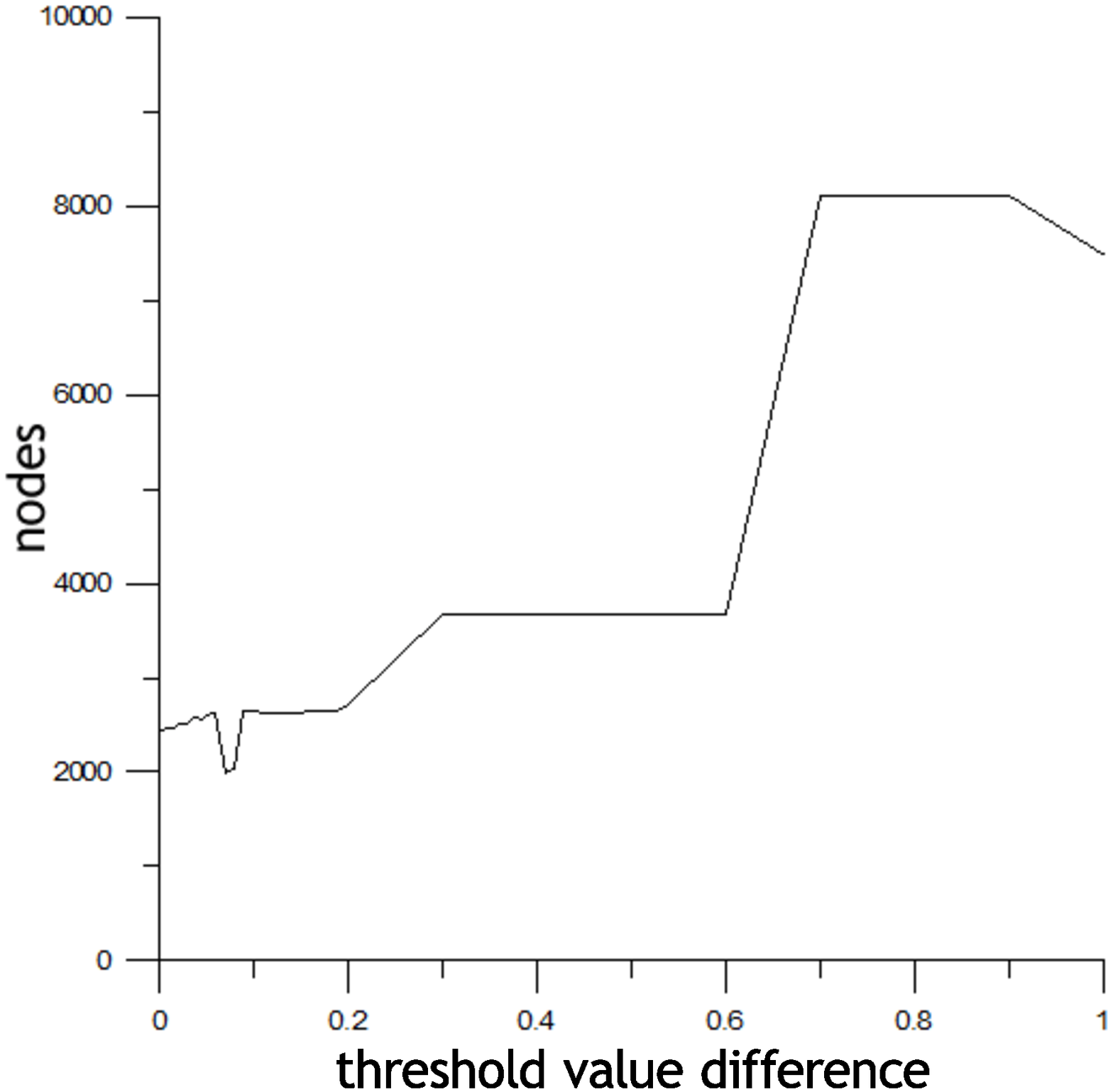}\\
    (a)&(b)\\
    \includegraphics[height=1.6in]{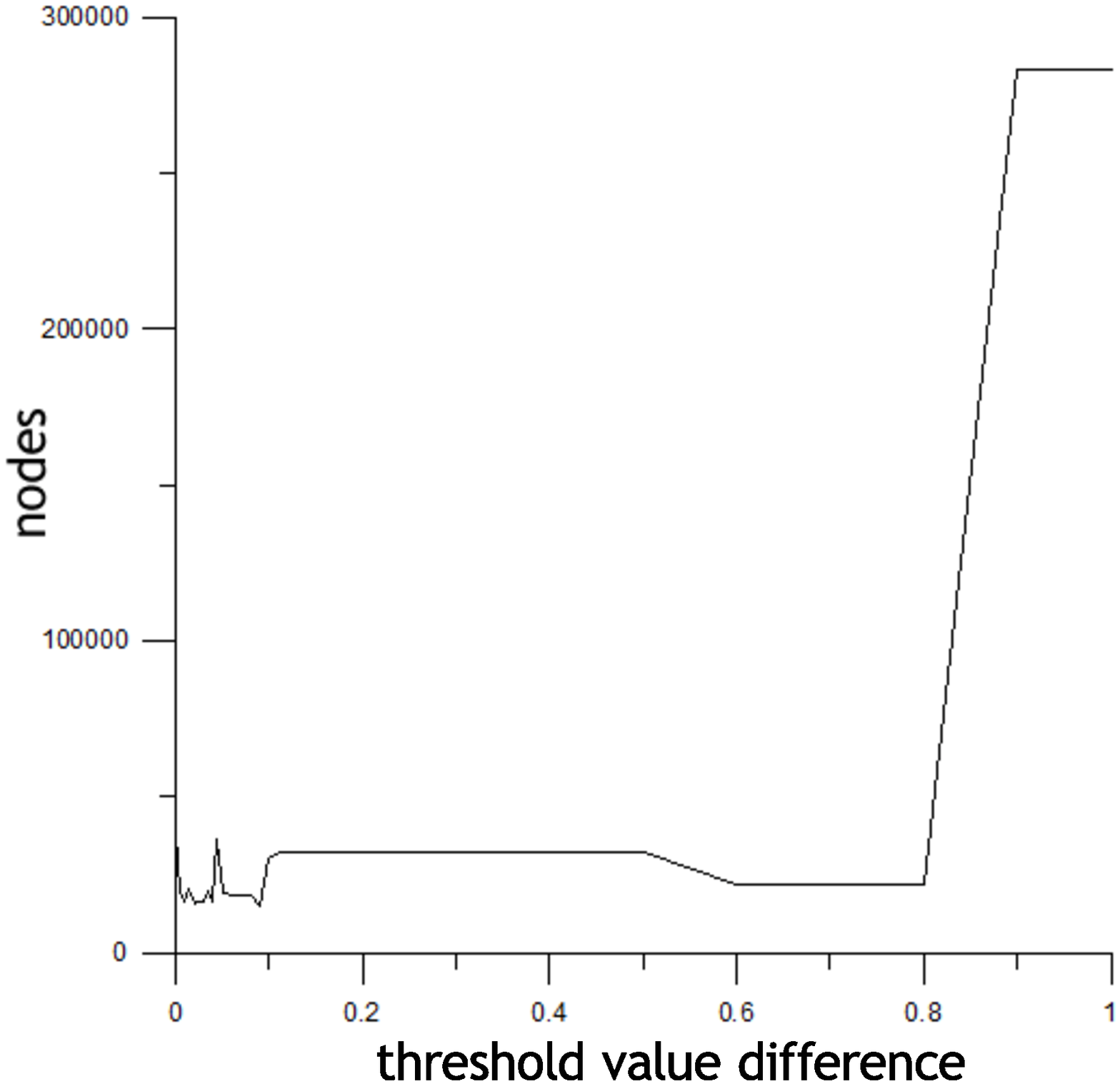} &
    \includegraphics[height=1.6in]{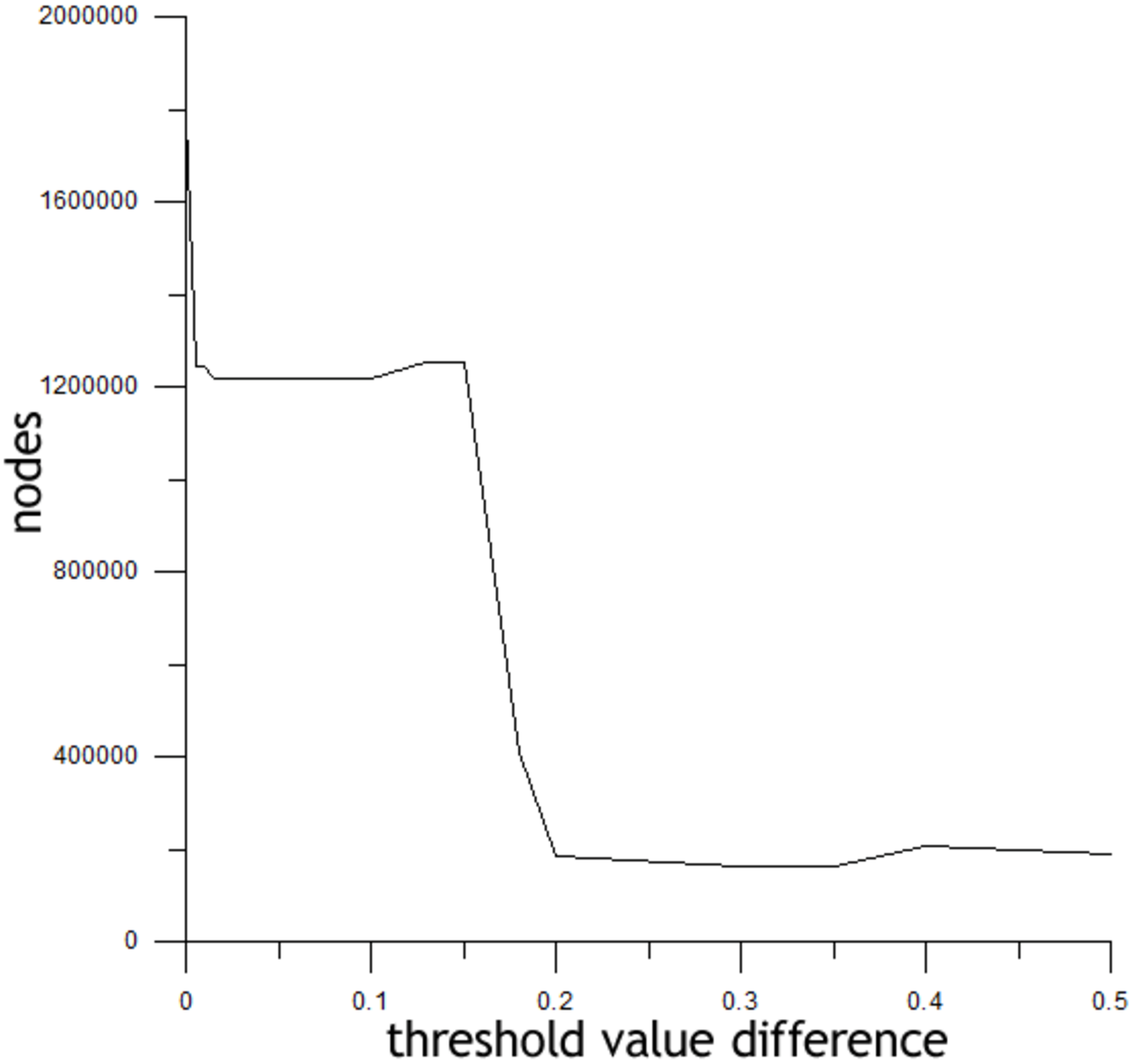}\\
    (c)&(d)
  \end{tabular}
\caption{(a) variable changes and (b) nodes over increasing values of $e$ for the scen11 RLFAP. (c), (d) nodes for the series12 (c) and haystacks-05 (d) instances.}
\label{fig:plot}
\end{figure}

Specifically for $dom/wdeg$, the value of $e$ was increased in steps of 0.01. Figures~\ref{fig:plot}b, c and d show the number of nodes ($y$-axis) as a relation of $e$ ($x$-axis) for instances scen11, series12 and haystacks-05 respectively. 
The first data point in these plots (where $e=0$) essentially gives the number of nodes for 2-way branching. The experiment was stopped when for some value of $e$ the observed value of $vc$ was 0. The last data point in each plot, corresponding to this situation, essentially gives the number of nodes for restricted 2-way branching.

In Figures~\ref{fig:plot}b and c (scen11 and series12), where 2-way branching is better than its restricted version, we can notice that as $e$ increases there is point where a sharp decline in the performance of $H_{sdiff}(e)$ occurs. Respectively, in Figure~\ref{fig:plot}d (haystacks-05) where restricted 2-way is better, we can notice that as $e$ increases there is point where a sharp improvement in the performance occurs. After running similar experiments with benchmark instances from other problem classes, we observed that setting $e$ to values around 0.1, when using $dom/wdeg$ for variable ordering, gives good results across many different instances and problem classes. For heuristic $dom/alldel$ this value was 0.001.

Finally, Figure~\ref{fig:plot}a gives the recorded value of $vc$ ($y$-axis) as a relation of $e$ ($x$-axis) for the RLFAP instance scen11. Not surprisingly, as the value of $e$ increases, the value of $vc$ decreases. Interestingly, there is a sharp (exponential) decline in the value of $vc$ roughly around the point where $e=0.1$. A similar phenomenon was observed in all the tested instances.

\subsection{Experimental Evaluation}


\begin{table*}[htb]
\caption{CPU times (t) in secs. and nodes (n) and variable changes (vc) for 2-way and the adaptive branching schemes using the {\em dom/wdeg} VOH.}
\centering
\begin{scriptsize}
\begin{tabular}{|c|c|c|c|c|c|c|c|}
\hline
Instance \T&  & $2-way$ & $restricted$ & $H_{sdiff}$ & $H_{cadv}$ & $H^{\wedge}$ & $H^{\vee}$\\
 &  &  & $2-way$ & (0.1) & $(wdeg)$ & & \\
\hline 
series-12 \T& t & 27.3 & 145.7 & 21 & \textbf{10.4} & 22.1 & 15.6\\
$(sat)$ & n & 33876 & 0.24M & 25730 & 13584 & 27721 & 17434\\
 & vc & 531 & 0 & 4 & 78 & 2 & 127\\ \hline
series-13 \T& t & 148.4 & 1132 & 98.3 & 150.6 & 70.6 & \textbf{59.1}\\
$(sat)$ & n & 0.15M & 1.6M & 96248 & 0.15M & 71570 & 64670\\
 & vc & 2492 & 0 & 9 & 1163 & 6 & 358\\ \hline
scen11 \T& t & \textbf{12.6} & 45.1 & 14 & 13 & 14 & 13\\
$(sat)$ & n & 2414 & 6002 & 2197 & 2456 & 2197 & 2023\\
 & vc & 31 & 0 & 8 & 31 & 8 & 31\\ \hline
scen2-f25 \T& t & 42.3 & 183.8 & 45 & \textbf{41.2} & 45 & \textbf{41.2}\\
$(unsat)$ & n & 9539 & 38185 & 9952 & 9561 & 9952 & 9561\\
 & vc & 719 & 0 & 69 & 484 & 69 & 484\\ \hline
graph9-f9 \T& t & \textbf{178.6} & 640.5 & 190 & 186.2 & 194.4 & 182\\
$(unsat)$ & n & 66367 & 0.25M & 69699 & 67585 & 70748 & 66802\\
 & vc & 873 & 0 & 74 & 698 & 68 & 700\\ \hline
ruler-25-8-a3 \T& t & \textbf{50.3} & 180.8 & 139.5 & 54.6 & 139.5 & 54.6\\
$(unsat)$ & n & 1829 & 6407 & 5022 & 1905 & 5022 & 1905\\
 & vc & 281 & 0 & 1 & 159 & 1 & 159\\ \hline
 frb35-17-1 \T& t & \textbf{34.2} & 135 & 69.4 & 41.8 & 66.4 & 41.8\\
$(sat)$ & n & 11129 & 54026 & 25003 & 15163 & 25003 & 15163\\
 & vc & 63 & 0 & 5 & 43 & 5 & 43\\ \hline 
\end{tabular}
\end{scriptsize}
\label{table:adaptiveWdeg} 
\end{table*}

We have experimented with 400 instances from the problem classes detailed in Section~\ref{section-comparison}. We compared the performance of the fixed branching schemes 2-way and restricted 2-way to the performance of the following adaptive branching schemes: $H_{sdiff}(0.1)$, $H_{cadv}(wdeg)$, and their conjunctive and disjunctive combinations $H^{\wedge}$  and $H^{\vee}$. The value of the threshold $e$ for $H_{sdiff}$ was set to 0.1 after a few preliminary experiments as explained above, while the secondary heuristic used by $H_{cadv}$ was $wdeg$. The VOH used was {\em dom/wdeg}. 

In 51\% of the tried instances all methods were very close. In 24\% 2-way was faster than restricted 2-way, while in 25\% it was slower. The last two cases included most of the hardest instances. Tables~\ref{table:adaptiveWdeg} and~\ref{table:adaptiveWdeg1} give some results from the last two cases indicating the gaps in performance between 2-way and restricted 2-way and the fact that the adaptive heuristics typically follow the winner between the two branching schemes. For each instance and branching scheme we report CPU time, nodes, and the observed value of $vc$. 

Table~\ref{table:adaptiveWdeg} includes instances where 2-way is better than restricted 2-way, while Table~\ref{table:adaptiveWdeg1} includes instances where restricted 2-way is better than 2-way. The adaptive branching schemes are, in most cases, close to or even slightly superior to 2-way branching in Table~\ref{table:adaptiveWdeg}, while a similar observation can be made with respect to restricted 2-way branching for Table~\ref{table:adaptiveWdeg1}. Respectively, the adaptive branching schemes are clearly superior to restricted 2-way in the Table~\ref{table:adaptiveWdeg}, and to 2-way in Table~\ref{table:adaptiveWdeg1}.
Generally, we can notice that although the adaptive branching schemes do not always achieve the best performance, they obtain a good trade-off between the performance of 2-way and restricted 2-way branching. Additionally, in many instances the adaptive branching schemes are superior to both 2-way and restricted 2-way (e.g. series-12 and frb35-17-2). Note that the tables include instances where one or both of the adaptive methods performed substantially worse than the winner among the standard branching schemes (e.g. ruler-25-8-a3 and haystacks-05). 

Comparing the heuristics, $H_{sdiff}$ and $H^{\wedge}$ are more efficient and robust than $H_{cadv}$ and $H^{\vee}$. However, we have not experimented extensively with secondary advisors for $H_{cadv}$ so far. The one used in the experiments ({\em wdeg}) is obviously closely related to the primary VOH ({\em dom/wdeg}). Trying a more diverse secondary advisor, i.e. a heuristic that utilizes different type of information, may yield better results. Note that using {\em dom} as the secondary advisor is clearly inferior compared to {\em wdeg} (results are omitted due to lack of space). This is not surprising if we consider that {\em wdeg} is a much more effective VOH compared to {\em dom}.

Taking a closer look at the results presented in
Table~\ref{table:adaptiveWdeg} it is interesting to notice the behavior of the branching schemes on the first instance series12. Here, restricted 2-way is clearly inefficient compared to 2-way. The latter branches on a variable different than the current one after a right branch 531 times throughout search (i.e. $vc=531$). On the other hand, $H_{sdiff}$ and $H^{\wedge}$ manage to outperform 2-way branching by only branching on 4 (resp. 2) different variables after right branches. Restricted 2-way branching is outperformed by a factor of 7, by only making 4 (resp. 2) decisions against the VOH. Similar behavior can be observed in most instances of Table~\ref{table:adaptiveWdeg}. Even more so in Table~\ref{table:adaptiveWdeg1} where restricted 2-way is better that 2-way, $H_{sdiff}$ and $H^{\wedge}$ follow the VOH very few times when it chooses to branch on a variable other than the current one.

\begin{table*}[hbt]
\caption{CPU times (t) in secs. and nodes (n) and variable changes (vc) for 2-way and the adaptive branching schemes using the {\em dom/wdeg} VOH.}
\centering
\begin{scriptsize}
\begin{tabular}{|c|c|c|c|c|c|c|c|}
\hline
Instance \T&  & $2-way$ & $restricted$ & $H_{sdiff}$ & $H_{cadv}$ & $H^{\wedge}$ & $H^{\vee}$\\
 &  &  & $2-way$ & (0.1) & $(wdeg)$ & & \\
\hline 
geo-d4-75-1 \T& t & 2298 & 1242 & \textbf{1233} & 2311 & \textbf{1233} & 2313\\
$(sat)$ & n & 0.55M & 0.28M & 0.27M & 0.45M & 0.27M & 0.45M\\
 & vc & 1428 & 0 & 2 & 689 & 2 & 690\\ \hline
geo-d4-75-2 \T& t & 122 & \textbf{28.5} & 37 & 57 & 37 & 57\\
$(sat)$ & n & 48943 & 9153 & 12356 & 18995 & 12356 & 18995\\
 & vc & 642 & 0 & 2 & 58 & 2 & 58\\ \hline
haystacks-05 \T& t & 41.7 & \textbf{3.9} & 28.1 & 41.7 & 28.1 & 41.7\\
$(unsat)$ & n & 1.78M & 0.18M & 1.2M & 1.78M & 1.2M & 1.78M\\
 & vc & 20 & 0 & 7 & 20 & 7 & 20\\ \hline
qcp-15-120-5 \T& t & 529 & \textbf{408} & 409 & 418 & 409 & 418\\
$(sat)$ & n & 0.76M & 0.6M & 0.6M & 0.61M & 0.6M & 0.61M\\
 & vc & 3177 & 0 & 1 & 1347 & 1 & 1347 \\ \hline
abb313-7 \T& t & 498 & \textbf{313} & \textbf{313} & 429 & 313 & 429\\
$(unsat)$ & n & 28388 & 17238 & 17238 & 21699 & 17238 & 21699\\
 & vc & 6 & 0 & 0 & 4 & 0 & 4\\ \hline
ruler-25-7-a3 \T& t & 12.4 & 4.1 & 5.8 & \textbf{3} & 5.8 & \textbf{3}\\
$(sat)$ & n & 1444 & 255 & 291 & 190 & 291 & 190\\
 & vc & 28 & 0 & 3 & 7 & 3 & 7\\ \hline
frb35-17-2 \T& t & 337 & 227 & 124 & \textbf{99.1} & 113 & \textbf{99.1}\\
$(sat)$ & n & 0.13M & 0.1M & 56508 & 42712 & 46935 & 42712\\
 & vc & 782 & 0 & 6 & 120 & 5 & 120\\ \hline
\end{tabular}
\end{scriptsize}
\label{table:adaptiveWdeg1} 
\end{table*}

These results suggest that heuristic $H_{sdiff}$ in particular is able to ``block'' variable changes that have a degrading effect on the search effort. Heuristic $H_{cadv}$ also achieves this, but to a lesser extent, as is evident by the $vc$ numbers. The behavior of the $H^{\wedge}$ (resp. $H^{\vee}$) heuristic is closely related to that of $H_{sdiff}$ (resp. $H_{cadv}$).

To verify these, we rerun all the experiments and each time a different variable $y$ than the current one $x$ was selected at a right branch, we ordered all the 
variables according to their {\em dom/wdeg} value. Then we measured the distance ($dis$) between $x$ and $y$ in this ordering (obviously $y$ was always first). 
Results show that the average value of $dis$ for $H_{sdiff}$ was significantly larger than the average $dis$ for 2-way branching. For example in the series12 instance, average $dis$ for $H_{sdiff}$ was 9 while for 2-way branching it was 1.9. For the geo50-20-d4-75-1 instance average $dis$ for $H_{sdiff}$ and 2-way was 14 and 1.6 respectively. These results demonstrate that  $H_{sdiff}$ allows variable changes only when the selected variable is considerably superior to the current variable according to the VOH.

Finally, in order to evaluate the statistical significance of our experimental results, a statistical analysis through a set of \emph{paired t-tests} was performed. We analyzed the CPU performance of the adaptive branching schemes compared to 2-way and restricted 2-way, over the 400 instances. We measured the mean difference in secs., standard deviation, t-value and the 95\% confidence interval. The risk level (called alpha level) was set to 0.05. Results are collected in Table~\ref{table:ttest}. The mean CPU reduction is always greater than zero. However, the negative values at the confidence interval indicates that this reduction was not observed in all the tried instances. According to standard tables of significance (available as an appendix in the back of most statistics texts) we can confirm that t-values for $H_{sdiff}$ were large enough to be significant. This is not clear for $H_{cadv}$, which means that the observed improvement is not large enough to be clearly significant.

We have also run experiments with the conflict driven VOH $dom/alldel$ (not reported here due to lack of space) in which adaptive branching schemes with heuristics $H_{sdiff}(0.001)$ and $H_{cadv}(wdeg)$ were again on average more efficient than the fixed branching schemes.

\begin{table}
\caption{Paired t-test measurements for evaluation of the significance of the experimental results.}
\centering
\begin{scriptsize}
\begin{tabular}{|c|c|c|c|c|c|}
\hline
 & Mean & SD & t-value  & 95\% C.I.\\ \hline
2-way vs $H_{sdiff}$ & 61.02 & 232.7 & 2.966 & (18.96, 103.09) \\ \hline
restr. 2-way vs $H_{sdiff}$  & 332.2 & 2956 & 1.231 & (-202, 866)\\ \hline
2-way vs $H_{cadv}$  & 8.03 & 135.3 & 0.65 & (-16.3, 32.5)\\ \hline
restr. 2-way vs $H_{cadv}$ & 279.2 & 2982 & 1.025 & (-259, 818) \\ \hline
\end{tabular}
\end{scriptsize}
\label{table:ttest} 
\end{table}

\section{CONCLUSIONS}


We compared the two most widely used branching schemes for CSPs, 2-way and $d$-way branching. Results showed that the theoretical benefits of 2-way branching are confirmed in practice once we move from a simple VOH like {\em dom} to a more sophisticated one like {\em dom/ddeg}. However, perhaps unexpectedly, when a state-of-the-art heuristic like {\em dom/wdeg} is used then there exist many cases where 2-way branching is significantly inferior to $d$-way and a restricted version of 2-way that is commonly used.

We then introduced generic heuristics that can be used to dynamically decide whether the VOH will be followed or not at certain points during search. The application of such heuristics results in an adaptive branching scheme that switches between 2-way branching and its restricted version, which is close to $d$-way branching. Experiments with instantiations of the generic heuristics confirm that search with adaptive branching outperforms search with a fixed branching scheme on a wide range of problems.

The work presented here is, to the best of our knowledge, the first attempt towards designing heuristics for adaptive branching and contributes to the design and implementation of adaptive constraint solvers.


\label{section-conclusions}

\bibliography{extra}
\end{document}